# A Bi-LSTM Autoencoder Framework for Anomaly Detection – A Case Study of a Wind Power Dataset

Ahmed Shoyeb Raihan[1] and Imtiaz Ahmed[2*]

*Abstract*— Anomalies refer to data points or events that deviate from normal and homogeneous events, which can include fraudulent activities, network infiltrations, equipment malfunctions, process changes, or other significant but infrequent events. Prompt detection of such events can prevent potential losses in terms of finances, information, and human resources. With the advancement of computational capabilities and the availability of large datasets, anomaly detection has become a major area of research. Among these, anomaly detection in time series has gained more attention recently due to the added complexity imposed by the time dimension. This study presents a novel framework for time series anomaly detection using a combination of Bidirectional Long Short Term Memory (Bi-LSTM) architecture and Autoencoder. The Bi-LSTM network, which comprises two unidirectional LSTM networks, can analyze the time series data from both directions and thus effectively discover the long-term dependencies hidden in the sequential data. Meanwhile, the Autoencoder mechanism helps to establish the optimal threshold beyond which an event can be classified as an anomaly. To demonstrate the effectiveness of the proposed framework, it is applied to a real-world multivariate time series dataset collected from a wind farm. The Bi-LSTM Autoencoder model achieved a classification accuracy of 96.79% and outperformed more commonly used LSTM Autoencoder models.

## I. INTRODUCTION

Anomalies are data points or groups of data points that exhibit significant deviation from their neighboring points or clusters and do not conform to the overall trend of the data [1]. Anomaly detection refers to the process of identifying uncommon occurrences, objects, or observations that deviate substantially from typical behaviors or patterns [2]. Anomaly detection has gained significant attention in various research fields, such as machine learning and data mining, networking, health, cybersecurity, fraud detection, and industrial process control. This is primarily because unusual events can offer valuable insights into the phenomenon being studied [3]. With the advent of Industry 4.0 technologies and the increasing availability of big datasets, scientists and researchers are relying more on data-driven approaches that leverage machine learning frameworks to address various problems [4]. Techniques like deep learning and sequential learning can produce superior results with less time and resource consumption, as they rely solely on the available data [5]. However, datasets can often contain anomalies or corrupted information that may result in inaccurate or inefficient prediction and classification models. Therefore, detecting anomalies and eliminating them from datasets before using them in ML-based models is a critical step. Traditional techniques for detecting anomalies rely on statistical and time-invariant methods, which are insufficient in dealing with the intricate and ever-changing nature of anomalies. The majority of present research on anomaly detection focuses solely on learning typical and anomalous behaviors and fails to incorporate temporal patterns when identifying new incoming anomalies [6]. However, due to the growing significance of anomaly detection and prevention in various fields and the progress of artificial intelligence, neural network approaches are often used these days which can identify more sophisticated types of anomalies and account for temporal and contextual attributes [7]. Long Short-Term Memory (LSTM) is one of the most popular recurrent neural network (RNN) architectures that can effectively represent the connection between present and past events, while also learning the long-term dependencies of a time-series dataset [8]. LSTM is often viewed as an extension of a recurrent architecture. RNNs are capable of modeling sequential data by utilizing information from the preceding time step, however, they encounter the vanishing gradient problem while attempting to capture long-term dependencies. On the contrary, LSTMs are specifically created to solve this issue by utilizing a memory cell that can choose to include or exclude information, thereby enhancing their ability to manage long-term dependencies [9]. Although more efficient than RNN, the unidirectional LSTM approach suffers in presence of long and complex sequences as it process sequence from one direction only [10]–[12]. In this regard, a Bi-LSTM-based approach employs two unidirectional LSTM networks to process temporal data. One LSTM network processes data from past to future, while the other processes data from future to past. This approach enhances the efficiency of the feature extraction process from temporal data, resulting in better anomaly detection performance [13].

In this work, we have employed an integrated deep learning approach combining a Bi-LSTM neural network and an Autoencoder to detect the anomalies in a wind power dataset. While the Bi-LSTM network effectively learns the time-related information inherent in the multivariate timeseries dataset, the autoencoder helps to set the threshold for detecting the anomalies based on the reconstruction error. The rest of the paper unfolds as follows. Section II discusses the wind power dataset. Section III elaborates on the main components of our model. Section IV describes the model architecture explaining the anomaly detection mechanism. The proposed model is implemented on the wind power dataset in section V and the results are discussed in section VI. Finally, the paper is concluded in section VII.

*Corresponding author.

[1]Ahmed Shoyeb Raihan is with the Department of Industrial & Management Systems Engineering, West Virginia University, WV, USA. ( e-mail: ar00065@mix.wvu.edu).

[2]Imtiaz Ahmed is with the Department of Industrial & Management Systems Engineering, West Virginia University, USA. (phone: 979-985-7439; e-mail: imtiaz.ahmed@mail.wvu.edu).

## II. DATASET

The wind power dataset is obtained from a wind farm, which includes 30,997 data records and covers the entire year of 2015. Wind farm operation data are recorded as a ten-minute average. This is a multivariate time-series dataset with four environmental variables or features. The first feature is the average wind speed (V), the second feature is the standard deviation of wind speed ($V_s$), the third feature is the average wind direction (D) in degrees, (from 0 to 360, where 0 degree means north), and the fourth and last feature is the temperature (T) expressed in degree Celsius. The target response is the wind power (y) measured on the turbine, expressed in a value between 0 and 1. Figure 1 illustrates the pattern of these four features for the first 20000 timestamps.

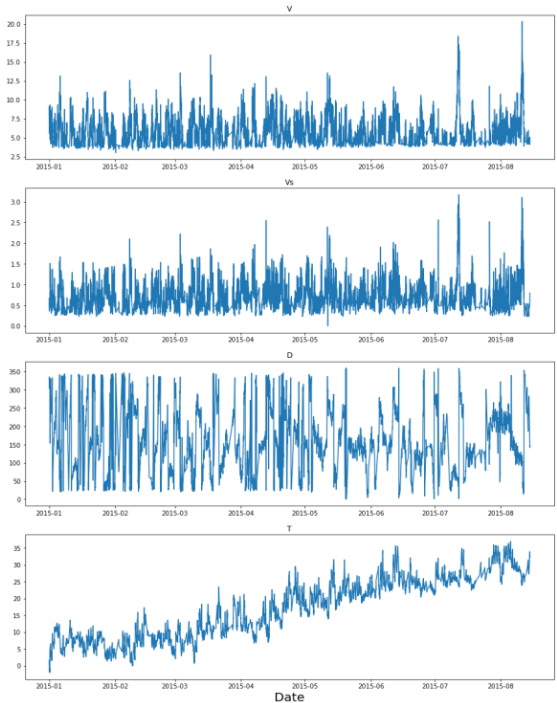

Figure 1. Visualization of the four features of the wind power data

## III. BI-LSTM AUTOENCODER

A Bi-LSTM Autoencoder model for anomaly detection composed of a neural network architecture that uses Bi-LSTM units to encode and decode input timeseries data. The goal of the model is to learn a compressed representation of the input data in the encoding phase and then reconstruct the original timeseries data in the decoding phase. The model consists of two main components: the encoder and the decoder. The encoder takes the input timeseries data and maps it to a lower-dimensional representation using Bi-LSTM layers. The decoder then takes the encoded representation and reconstructs the original timeseries data using another set of Bi-LSTM layers. To detect anomalies, the reconstruction error between the input timeseries data and the output of the decoder is measured. Anomalies are identified as data points with high reconstruction errors. This is because the model has learned the normal data patterns and any deviations from these patterns will result in high reconstruction errors. The Bi-LSTM autoencoder model is trained on a dataset consisting of normal data only. During inference, the model reconstructs new timeseries data and identifies any anomalies based on the reconstruction error. LSTM-based architectures have been extensively used in anomaly detection of timeseries data such as Stacked-LSTM [9], Stacked LSTM-SVM [14], Contractive LSTM [15], Variational LSTM AE [16], Observer-based LSTM AE [17] and CNN-based LSTM [18]. By using a Bi-LSTM Autoencoder-based approach we can enhance the accuracy of predictions by training the network not only on past data to forecast the future but also on future data to forecast the past. The components of a Bi-LSTM Autoencoder are explained in the following subsections.

### A. LSTM

An LSTM cell is the building block of our proposed model. LSTM is considered an upgraded form of RNN. One of the main limitations of RNNs is the vanishing gradient problem, which occurs when the gradients of the network weights become too small during backpropagation, causing the network to stop learning. LSTM networks solve this problem by introducing gates that regulate the flow of information in the network, allowing it to selectively remember or forget information over long periods. Another advantage of LSTM networks is their ability to capture long-term dependencies in sequential data. Unlike RNNs, which may forget important information over long sequences, LSTMs can remember important information for much longer periods, making them more effective at capturing complex patterns and relationships in sequential data. Figure 2 shows the components of a typical LSTM unit, which consists of a cell, an input gate, an output gate, and a forget gate. The three gates control the flow of information into and out of the cell, and the cell remembers values across arbitrary time intervals. The detailed working mechanism of an LSTM unit can be explained through the elaboration of its three gates: forget gate, input gate and output gate [19].

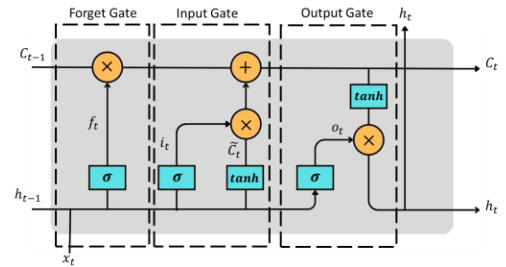

Figure 2. An LSTM cell with three gates

- Forget Gate: The forget gate's primary function is to determine which bits of the cell state are helpful. To achieve this, the neural network is fed both the old hidden state and the fresh input data. Using a sigmoid activation function, the neural network creates a vector with each element falling within the range of [0,1]. The output denoted by $f_t$ from the forget gate can be expressed using Equation 1. In this equation, σ is the activation function, $w_f$ and $b_f$ are the weight and bias of the forget gate, respectively. $H_{t-1}$ and $X_t$ represent the concatenation of the last hidden state and the current input, respectively.

$$f_t = \sigma(w_f[H_{t-1}, X_t] + b_f) \quad (1)$$

- Input Gate: The input gate has two primary goals. The first step is to determine what part of the new data should be used for the cell/memory update. Second, it tries to find out whether the new input data is worth remembering at all. The input gate undergoes two stages to achieve this. These two stages are depicted using the following Equations (2) and (3). In Equation (2), the weight matrices and bias of the operation are represented by $w_c$ and $b_c$, respectively. Meanwhile, the activation function used is the hyperbolic tangent ($tanh$) function. In equation (3), $w_i$ and $b_i$ are the weight matrices and the bias of the input gate, respectively.

$$\tilde{C}_t = tanh(w_c[H_{t-1}, X_t] + b_c) \quad (2)$$
$$i_t = \sigma(w_i[H_{t-1}, X_t] + b_i) \quad (3)$$

Pointwise multiplication is then used to update the cell state given by Equation 4.

$$C_t = f_t \odot C_{t-1} + i_t \odot \tilde{C}_t \quad (4)$$

- Output Gate: Determining the new hidden state is the output gate's primary function. To produce the filter vector $o_t$ as given in Equation 5, the prior hidden state and present input data are passed through the sigmoid-activated network.

$$o_t = \sigma(w_o[H_{t-1}, X_t] + b_o) \quad (5)$$

The filter vector is then applied to the cell state by pointwise multiplication after being passed through a $tanh$ activation function that compresses the values into the range [-1, 1]. Equation 6 illustrates the creation and output of a new hidden state $H_t$ along with a new cell state $C_t$.

$$H_t = o_t \odot tanh(C_t) \quad (6)$$

The new hidden state $H_t$ becomes a prior hidden state $H_{t-1}$ to the following LSTM unit, whereas the new cell state $C_t$ becomes a previous cell state $C_{t-1}$.

### B. Bi-LSTM

An LSTM layer is made up of a sequence of LSTM cells, and the input data is processed only in a forward direction. On the other hand, Bi-LSTM includes an additional LSTM layer that processes the data in a backward direction, as shown in Figure 3. Training a Bi-LSTM network is equivalent to training two separate unidirectional LSTM networks. One of these networks is trained on the original input sequence, while the other is trained on a reversed copy of the input sequence. This approach provides the network with more contextual information, leading to faster and more comprehensive learning of the problem.

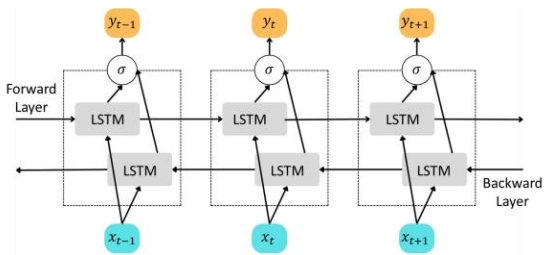

Figure 3: An unfolded view of Bi-LSTM

### C. Autoencoder

Autoencoders are a neural network architecture utilized for unsupervised learning, data compression, dimensionality reduction, and data generation. The fundamental concept behind autoencoders involves acquiring a compressed representation of the input data (encoding) that can later be utilized to rebuild the original data (decoding) with minimum loss of information. An input layer, an output layer, and multiple hidden layers make up a conventional Autoencoder. A simple Autoencoder model is shown in Figure 4 which can be explained by the encoder part, decoder part, and the reconstruction error (loss) between the input and output data [19].

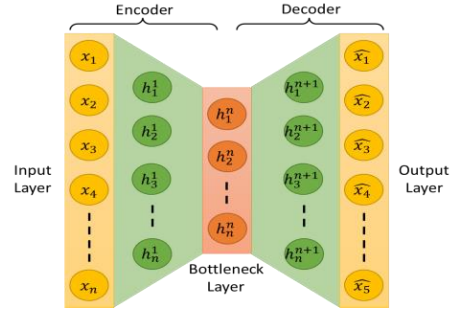

Figure 4: A simple Autoencoder architecture

- Encoding: The encoding operation maps the input data x, which is a high-dimensional vector ($x \in \mathbb{R}^m$), to a low-dimensional bottleneck layer representation (h). This is given by the following Equation where $w_i$ is the weight matrix, $b_i$ is a bias and $f_1$ is an activation function.

$$h = f_1(w_i x + b_i) \quad (7)$$

- Decoding: Equation 8 illustrates how the decoding operation uses the bottleneck layer representation of (h) to produce the output $\hat{x}$ which attempts to reconstruct $x$. Here, $f_2$ is an activation function for the decoder, $w_j$ is the weight matrix, $b_j$ represents a bias and $\hat{x}$ represent reconstructed input sample.

$$\hat{x} = f_2(w_j h + b_j) \quad (8)$$

- Reconstruction Loss: As illustrated in Equation 9, a reconstruction loss ($L$) is calculated in a typical Autoencoder model to minimize the difference between the output and the input. Here, $x$ represents the input data, $\hat{x}$ indicates the output/predicted data, and $n$ is the number of samples in the training dataset. This reconstruction loss is frequently employed in tasks involving anomaly identification.

$$\min_{w,b} L(x - \hat{x}); where\ L(x - \hat{x}) = \frac{1}{n}\sum_{n=1}^{n}|\hat{x}_t - x_t| \quad (9)$$

## IV. METHODOLOGY

In this section, we present our suggested framework, which analyzes timeseries data to identify anomalies using an Autoencoder equipped with a Bi-LSTM architecture. Our proposed model architecture is illustrated in Figure 5. The architecture is composed of four stages: a timeseries input, a Bi-LSTM encoder, a Bi-LSTM decoder, and an anomaly

detection module. These stages are discussed in the following:

- Timeseries Data Input: The original wind power dataset, at first, needs to be converted to a suitable format to pass as an input to the LSTM cells. The inputs to LSTMs are 3-dimensional arrays created from the timeseries data. The shape of such an array is $samples \times lookback \times features$. Here, samples are the number of observations, or in other words, the number of data points. In our dataset, there is a total of 30997 samples for training and testing purposes. LSTM models are meant to look at the past which suggests that at time $t$, the LSTM will process data up to $(t - lookback)$ to make a prediction. Features are the number of variables present in the timeseries data. Our dataset has four features $V$, $V_s$, $D$ and $T$. After processing the original dataset, we have $[X_1, X_2, X_3, ..., X_n]$ samples where $n = 30997$. Each sample $X_i$ is thus a 2-dimensional array of shapes equivalent to $lookback \times features$, constructed with a lookback period of $T$ and data $[x_1, x_2, x_3, ... x_t]$, where $x_t \in \mathbb{R}^m$, denotes an input with $m$ features at time instance $t$.

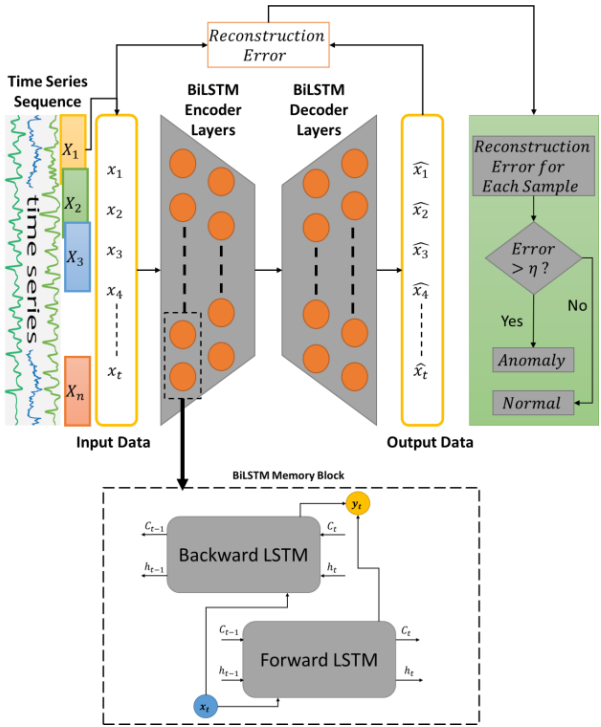

Figure 5: Proposed Bi-LSTM Autoencoder architecture for anomaly detection

- Bi-LSTM Encoder: The encoder takes a sequence of high-dimensional input data and produces a fixed-size vector. By utilizing the memory cells of Bi-LSTM, the encoder preserves dependencies across multiple data points in a time-series sequence with more efficiency while simultaneously reducing the high-dimensional input vector representation to a lower-dimensional representation. In our model, the Bi-LSTM encoder has two layers. The first layer has 64 LSTM cells whereas the second layer has 32 LSTM cells. The output of this second layer is the encoded feature vector of the input data.

- Bi-LSTM Decoder: The decoder LSTM generates a fixed-size input sequence from the reduced representation of the input data in the latent space. While reconstructing, it tries to minimize the difference between the original and reconstructed data i.e., reconstruction error. These reconstruction errors can be utilized to establish a threshold which can in turn identify anomalies. The input to the decoder is the output from the encoder in a reduced dimension. The decoder layer is designed to unfold the encoding. Therefore, the decoder layers are stacked in the reverse order of the encoder. The first decoder layer thus has 32 LSTM cells and the second decoder layer in our model has 64 LSTM cells. The final output from the decoder is the reconstruction of the input to our model.

- Anomaly Detection: An observation that deviates from the majority of the data might be referred to as an anomaly [19]. To determine how far an observation deviates, a threshold needs to be established. Anomalies are any observations that are outside of this threshold. Our model is trained on a dataset that comprises the response ($y$) values that are negatively labeled i.e., there are no anomalies present in the training dataset. It can provide us the reconstruction error values related to the normal data points only. Once training is complete and all distinct reconstruction errors have been computed, we can study the distribution of the reconstruction error values and establish a suitable threshold ($\eta$) for anomaly detection. After the threshold has been set, we pass the test data to the trained model. The test data includes both positively labeled and negatively labeled response values. For each sample in the testing set, a reconstruction error value is calculated. The sample is regarded as an anomaly if the reconstruction error exceeds the threshold ($\eta$) as in Equation 10.

$$X' = \begin{cases} X'_i \text{ is an anomaly, if } ltest_{arr}[i] > \eta \\ X'_i \text{ is normal, otherwise} \end{cases} \quad (10)$$

Here, $X'$ indicates a reconstructed time series, $X'_i$ is a data point contained in the time series, and $ltest_{arr}[i]$ is a result of a reconstruction loss function using Equation 9.

V. IMPLEMENTATION ON THE WIND POWER DATASET

In this work, the wind power dataset consisted of a total of 30997 samples or data points. 70% of the data is used for training and validating the model and the remaining 30% of the data is used for testing. From the training set, the data was further divided into two classes: one for the initial training and the other for the validation purpose. Furthermore, the model has been trained with only negatively labeled samples i.e., no anomalies are present in the training set. This allows us to work with a training set of 14021 samples. Each sample has a lookback period of 10. The shape of the training data is thus 14021×10×4. The shape of the test data is 9296×10×4, which indicates that we have a total of 9296 samples in the test set, each sample has a lookback period of 10 with 4 features. It is important to note that, in the case of the test set, the samples contain both positively and negatively labeled response values.

The encoder part of the network has two Bi-LSTM layers; the first layer has 64 LSTM cells whereas the second layer has 32 LSTM cells. For the decoder part, the first LSTM layer has 32 units and the second one has 64 LSTM units. The model is trained with a learning rate of 0.0001 and a batch size of 128. After the training is done, the learning curve for loss and accuracy with the training and validation dataset is shown in Figure 6.

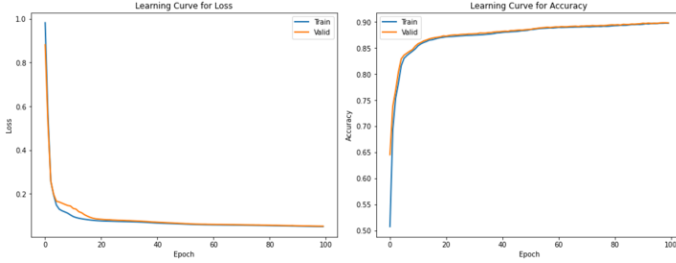

Figure 6: Learning curves for loss (left) and accuracy (right)

The training data is also used to calculate the reconstruction loss using Equation 9. A distribution of reconstruction loss of the training set which contains no anomalies is shown in Figure 7. The threshold beyond which a datapoint can be termed as an anomaly can be set using domain knowledge and this distribution.

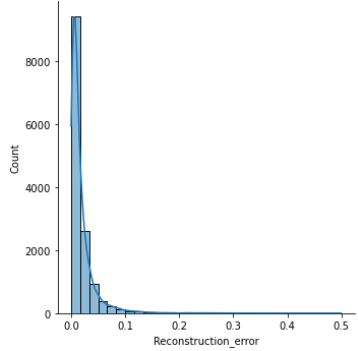

Figure 7: Distribution of the reconstruction loss of the training data

After training the model and setting the threshold to detect anomalies, the test dataset is now passed to the trained model. The reconstruction loss on the test dataset is visualized in Figure 8. We can see that the anomalous data points (orange dots) have a higher reconstruction error.

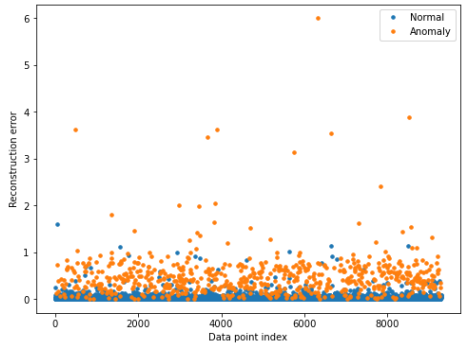

Figure 8: Reconstruction errors for normal and anomalous samples

## VI. RESULTS

We used some commonly used performance metrics to evaluate the performance of our proposed Bi-LSTM Autoencoder model on the wind power dataset. These metrics are accuracy, precision, recall, and F1 score. They are expressed using the following equations:

$$Accuracy = \frac{TP+TN}{TP+FP+TN+FN} \quad (11)$$

$$Precision = \frac{TP}{TP+FP} \quad (12)$$

$$Recall = \frac{TP}{TP+FN} \quad (13)$$

$$F1\ Score = 2 \times \frac{Precision \times Recall}{Precision+Recall} \quad (14)$$

Here, TP (True Positive) refers to the count of accurately identified anomalies, TN (True Negative) refers to the count of correctly identified normal events, FP (False Positive) indicates the count of normal events that were inaccurately diagnosed as anomalies, while FN (False Negative) denotes the count of anomalies that were incorrectly identified as normal events. A confusion matrix for the test dataset is illustrated in Figure 9.

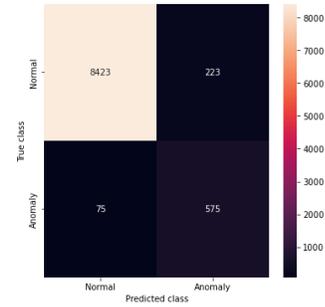

Figure 9: Confusion matrix of the test data

Using equations (11) - (14), we have also calculated the performance metrics for our proposed model. We also compared the performance of our model with the LSTM-based Autoencoder with similar architecture and parameters. The results are summarized in Table 1. The results show significant improvements in all the performance metrics by using Bi-LSTM Autoencoder.

TABLE I. Performance of Bi-LSTM Autoencoder

| Model | Accuracy | Precision | Recall | F1 Score |
|---|---|---|---|---|
| Bi-LSTM Autoencoder | 0.9679 | 0.7205 | 0.8846 | 0.7942 |
| LSTM Autoencoder | 0.9427 | 0.5598 | 0.8415 | 0.6724 |

The AUC-ROC graph in Figure 10 illustrates the effectiveness of the Bi-LSTM Autoencoder model in terms of the trade-off between true positive rate and false-positive rate. It achieved an AUC-ROC score of 0.97, which demonstrates that our proposed network is efficient at correctly identifying anomalies.

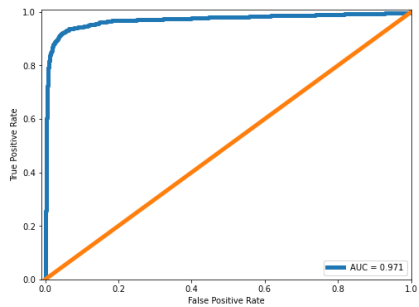

Figure 10: AUC-ROC curve of the proposed model

## VII. Conclusion

In this work, we have developed an anomaly detection framework equipped with Bi-LSTM architecture-based Autoencoder and demonstrated its application using a wind power dataset. The framework used four Bi-LSTM layers to analyze long-term relationships among time sequences more efficiently. The Autoencoder helps to reconstruct the data from latent space, captures the normal data distribution, and thereby detects abnormal events or anomalies. The proposed framework achieved promising results in the wind power dataset. Selecting a proper threshold for anomaly detection is one of the most critical steps of our framework. Choosing a smaller threshold value can increase the rate of anomaly detection, however, it can adversely affect the rate of false alarms. On the other hand, choosing a higher threshold value can result in too many miss detections while reducing the false alarm rate at the same time. In this work, we utilized both domain knowledge and the distribution of reconstruction errors to decide on the threshold for anomaly detection. We also compared the performance with LSTM Autoencoder which was comprehensively outperformed by our proposed model. We are currently working on integrating Variational Autoencoder (VAE) with Bi-LSTM to further improve the performance of the model.

## Acknowledgment

The authors are grateful for the continuous support of the Department of Industrial Management and Systems Engineering (IMSE) of West Virginia University for the completion of this work.